\documentclass[12pt,a4paper]{article}
\usepackage[utf8]{inputenc}
\usepackage[T1]{fontenc}
\usepackage{setspace}
\usepackage{geometry}
\usepackage{graphicx}
\usepackage{siunitx}
\usepackage{subcaption}
\usepackage[hidelinks]{hyperref}
\usepackage{titlesec}
\usepackage{booktabs}
\usepackage{array}
\usepackage{float}
\usepackage{amsmath}
\usepackage{minted}

\titleformat{\section}{\normalfont\Large\bfseries}{\thesection.}{1em}{}
\titleformat{\subsection}{\normalfont\large\bfseries}{\thesubsection.}{1em}{}
\titleformat{\subsubsection}{\normalfont\normalsize\bfseries}{\thesubsubsection.}{1em}{}

\title{\textbf{Graph Neural Network Assisted Genetic Algorithm for Structural Dynamic Response and Parameter Optimization}}

\author{
\textbf{Sagnik Mukherjee}$^{1}$, \textbf{Indrajit Barua}$^{2}$\\[1em]
\small
$^{1}$Department of Civil Engineering, Jadavpur University, Kolkata, India\\
\texttt{sagnikm.civil.ug@jadavpuruniversity.in}\\
$^{2}$Department of Civil Engineering, Jadavpur University, Kolkata, India\\
\texttt{ibarua.civil@jadavpuruniversity.in}
}

\begin{document}

\maketitle
\begin{abstract}
The optimization of structural parameters, such as mass ($m$), stiffness ($k$), and damping coefficient ($c$), is critical for designing efficient, resilient, and stable structures. Conventional numerical approaches, including Finite Element Method (FEM) and Computational Fluid Dynamics (CFD) simulations, provide high-fidelity results but are computationally expensive for iterative optimization tasks, as each evaluation requires solving the governing equations for every parameter combination. This study proposes a hybrid data-driven framework that integrates a Graph Neural Network (GNN) surrogate model with a Genetic Algorithm (GA) optimizer to overcome these challenges. The GNN is trained to accurately learn the nonlinear mapping between structural parameters and dynamic displacement responses, enabling rapid predictions without repeatedly solving the system equations. A dataset of single-degree-of-freedom (SDOF) system responses is generated using the Newmark–Beta method across diverse mass, stiffness, and damping configurations. The GA then searches for globally optimal parameter sets by minimizing predicted displacements and enhancing dynamic stability. Results demonstrate that the GNN–GA framework achieves strong convergence, robust generalization, and significantly reduced computational cost compared to conventional simulations. This approach highlights the effectiveness of combining machine learning surrogates with evolutionary optimization for automated and intelligent structural design.
\end{abstract}

\section{Introduction}
Optimizing structural parameters is critical since it affects efficiency, safety, sustainability, and reliability of engineering systems. It guarantees structures can meet the expectations for performance with arbitrary constraints with the least amount of material use, cost, and environmental footprint. 

For dynamic system performance, stiffness and damping parameters dictate vibration, stability, and fatigue life of structures. Optimizing these parameters enables engineers to reduce resonant response, improving structures' resistance to earthquakes or wind loads, and enhancing their service life.

Typically, Computerised Fluid Dynamics (CFD) based software, such as \textit{ANSYS} or \textit{COMSOL}
, or Finite Element Method (FEM) based numerical methodologies, are used to solve these types of optimization problems. However, regardless of the degree of accuracy the methodologies offer, they require excessive amounts of computational time—particularly in the scale of a large simulation just as it regularly involves repeated evaluations of parameters for optimization purposes. A traditional CFD or FEM based simulation requires a numerical solving approach to governing equations for each configuration, which depending on the system can equate to minutes to hours simulation time for complex systems.

In contrast, surrogate models produce instant predictions of the system responses once they are trained. This is extremely valuable when considering that one might want to evaluate hundreds if not thousands of combinations for possible parameters. This is especially valued in iterative optimization loops, or real-time systems.

Among these, \textbf{Graph Neural Networks (GNNs)} have demonstrated remarkable potential in the representation of physical systems with topological or relational dependencies. Unlike traditional neural networks that process vectorized data, GNNs can establish spatial and inter-element relationships among different structural parameters by representing or encoding them as graphs—where the nodes represent lumped masses or degrees of freedom, and edges signify stiffness or damping interactions. This graph representation allows GNNs to learn the dynamics of structural behavior directly from data and to predict structural responses such as displacement or maximum absolute displacement for the corresponding structural parametric combinations.

The optimization of structural parameters still remains a challenging task as the design space of mass (\(m\)), stiffness (\(k\)) and damping coefficient (\(c\)) is highly non-linear. To address this, metaheuristic algorithms that can iteratively calculate the optimum values can be incorporated. One such metaheuristic algorithm is the Genetic Algorithm (GA), which mimics the process of natural evolution. This algorithm provides a robust mechanism for global optimization, even in cases where gradient information is not available or unreliable, by iteratively evolving candidate solutions through selection, crossover, and mutation.
 Darwin formulated the fundamental principle of natural selection as the main evolutionary
principle long before the discovery of genetic mechanisms. Darwin hypothesized fusion or blending
inheritance, supposing that potential qualities were mixed together like fluids in the offspring organism.
Recently, genetic algorithms have received considerable attention regarding their potential as an
optimization technique for complex problems and have been successfully applied in the areas of industrial
engineering. GA is a stochastic search technique based on natural selection and genetics, developed by
Holland. This algorithm, differing from conventional search techniques, starting with the initial set of random solution called population. Each individual in a population is called a chromosome, representing a
solution to the problem at hand. A chromosome is a string of symbols; it is usually, but not necessarily, a
binary bit string. The chromosomes evolve through successive iterations, called generations. During each generation, the chromosomes are evaluated, using some measures of fitness. To create next generation,
new chromosomes, called offspring, are formed by either (a) merging two chromosomes form current
generation using cross over operator or (b) modifying a chromosome by mutation operator. A new
generation is formed by (a) selecting, according to fitness values, some of the parents and offspring and
(b) rejecting others so as to keep the population size constant. Fitter chromosomes have higher
probabilities of being selected. After several generations, the algorithms converge to the best chromosome,
which hopefully represents the optimal or suboptimal solution to the problem. 
Genetic Algorithms are easy to apply to a wide range of problems, from optimization problems like the
traveling salesperson problem, to inductive concept learning, scheduling, and layout problems. The results
can be very good on some problems, and rather poor on others. If only mutation is used, the algorithm is
very slow. Crossover makes the algorithm significantly faster. GA is a kind of hill-climbing search; more
specifically it is very similar to a randomized beam search. As with all hill-climbing algorithms, there is
a problem of local maxima. Local maxima in a genetic problem are those individuals that get stuck with
a pretty good, but not optimal, fitness measure. Any small mutation gives worse fitness. Fortunately,
crossover can help them get out of a local maximum. Also, mutation is a random process, so it is possible
that we may have a sudden large mutation to get these individuals out of this situation. (In fact, these
individuals never get out. It’s their offspring that get out of local maxima.) One significant difference
between GAs and hill-climbing is that, it is generally a good idea in GAs to fill the local maxima up with
individuals. Overall, GAs have less problems with local maxima than back-propagation neural networks.
If the conception of a computer algorithms being based on the evolutionary of organism is surprising,
the extensiveness with which this algorithms is applied in so many areas is no less than astonishing.
These applications, be they commercial, educational and scientific, are increasingly dependent on this
algorithms, the Genetic Algorithms. Its usefulness and gracefulness of solving problems has made it the a
more favorite choice among the traditional methods, namely gradient search, random search and others.
GAs are very helpful when the developer does not have precise domain expertise, because GAs possess
the ability to explore and learn from their domain.

\section{Objectives}

\subsection*{Primary Objective}
To develop a hybrid framework combining Graph Neural Networks (GNNs) and Genetic Algorithms (GAs) for optimizing structural parameters in dynamic systems.

\subsection*{Specific Objectives}

1) Model the Single Degree of Freedom (SDoF) system under varying vibration conditions.\\
2) Compare displacement time histories under different damping and excitation cases\\
3) Generate input data using the Newmark–Beta solver.\\
4) Design and implement a GNN–GA pipeline for prediction and optimization.\\
5) Predict the maximum absolute displacement for each (\(m, k, c\)) triplet.\\
6) Minimize displacement and determine optimal parameters.

\section{Methodology}
\section*{3.1: Modeling of Single Degree of Freedom (SDoF) Systems under Various Vibration Scenarios}

In the initial phase of this study, we focus on the dynamic modeling of Single Degree of Freedom (SDoF) systems subjected to diverse vibration inputs. The primary goal is to understand the system’s response characteristics across different structural configurations, defined by variations in mass ($m$), stiffness ($k$), and damping ($c$) properties.

The general equation for dynamic system establishing the relation between system parameters and displacement response is:

\begin{equation}
m \ddot{u}(t) + c \dot{u}(t) + k u(t) = F(t)
\end{equation}

where $F(t)$ represents the external excitation applied to the system. A systematic parametric approach is taken to simulate multiple SDoF systems under these excitation conditions, generating a comprehensive dataset of displacement time histories.

\subsection*{Time-History Plots}
Figure 1. illustrates sample time-history responses for selected SDoF configurations under different vibration scenarios. These plots provide insight into the dynamic behavior of the system under each excitation type.
For free vibration condition and under-damped system, the equation becomes:

\begin{equation}
m \ddot{u}(t) + c \dot{u}(t) + k u(t) = 0
\end{equation}

For a particular scenario under free-vibration where initial displacement is 0.01 m, \[
m = 1 kg,
k = 20 N/m,
\zeta = 0.05,
c = \zeta * 2 \sqrt{k m}
\]

\begin{figure}[!ht]
    \centering
    \includegraphics[width=0.8\textwidth]{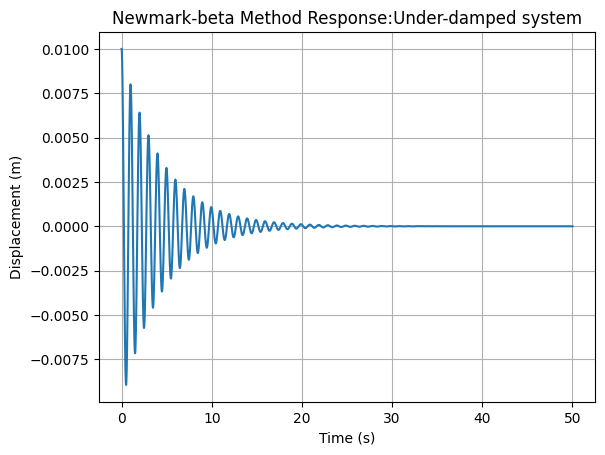} 
    \caption{Displacement time-history of SDoF under Free vibration condition}
    \label{fig:timehistory}
\end{figure}

\subsection*{Validation with Reference Data}
To ensure the accuracy and reliability of the simulations, selected cases are validated against published results, including tabulated data from Chopra's \textit{Dynamics of Structures}[1]. Table 1. and Fig. 2 present a comparison between the simulated peak displacements and reference values.

\begin{table}[htbp]
\centering
\caption{Numerical Solution by Newmark-Beta Method (Average Acceleration)}
\begin{tabular}{cccccc}
\hline
$t_i$ & $p_i$ & $u_i$ & $\dot{u}_i$ & $\ddot{u}_i$ \\
\hline
0.0 & 0.0000 & 0.0000 & 0.0000 & 0.0000 \\
0.1 & 5.0000 & 0.0439 & 0.8735 & 17.4666 \\
0.2 & 8.6602 & 0.2327 & 2.9056 & 23.1803 \\
0.3 & 10.0000 & 0.6121 & 4.6837 & 12.3724 \\
0.4 & 8.6603 & 1.0827 & 4.7262 & -11.5169 \\
0.5 & 5.0000 & 1.4309 & 2.2422 & -38.1611\\
0.6 & 0.0000 & 1.4231 & -2.3995 & -54.6733 \\
0.7 & -5.0000 & 0.9622 & -6.8183 & -33.7019 \\
0.8 & -8.6602 & 0.1906 & -8.6094 & -2.1229 \\
0.9 & -10.0000 & -0.6048 & -7.2936 & 28.4417 \\
1.0 & -8.2441 & -1.1444 & -3.5027 & 47.3716 \\
\hline
\end{tabular}

\end{table}
\begin{figure}[!ht]
    \centering
    \includegraphics[width=0.8\textwidth]{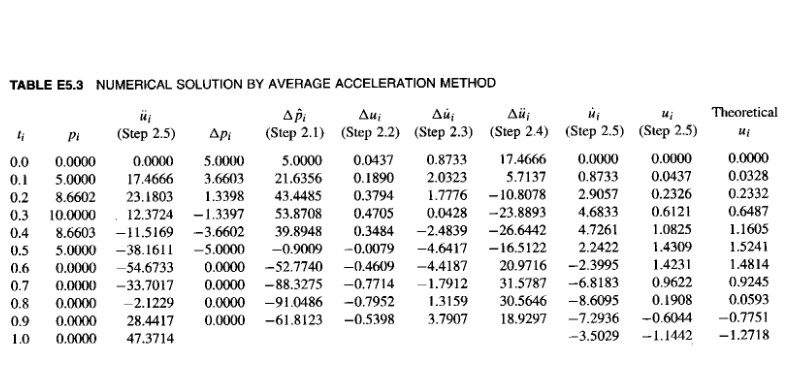} 
    \caption{Reference Data, Dynamics of structures by A.K.Chopra, Page 169}
    \label{output}
\end{figure}
\newpage
\thispagestyle{plain}
The equation of system under base excitation is:
\begin{equation}
m\ddot{u}(t) + c\dot{u}(t) + k u(t) = -m\ddot{u}_g(t)
\end{equation}\\
Relation Between Absolute and Relative Displacement:
\begin{equation}
\ddot{x}(t) = \ddot{u}(t) + \ddot{u}_g(t)
\end{equation}
Simplifying after substitution gives:
\begin{equation}
m\ddot{u}(t) + c\dot{u}(t) + k u(t) = -m\ddot{u}_g(t)
\end{equation}
For base excitation condition modelling, we have taken the 1940 Elcentro North-South Earthquake ground acceleration data~\hyperref[Elcentro_NS]{[2]}and modelled using Newmark Beta(Average acceleration) solver:
\begin{figure}[!ht]
    \centering
    \includegraphics[width=0.8\textwidth]{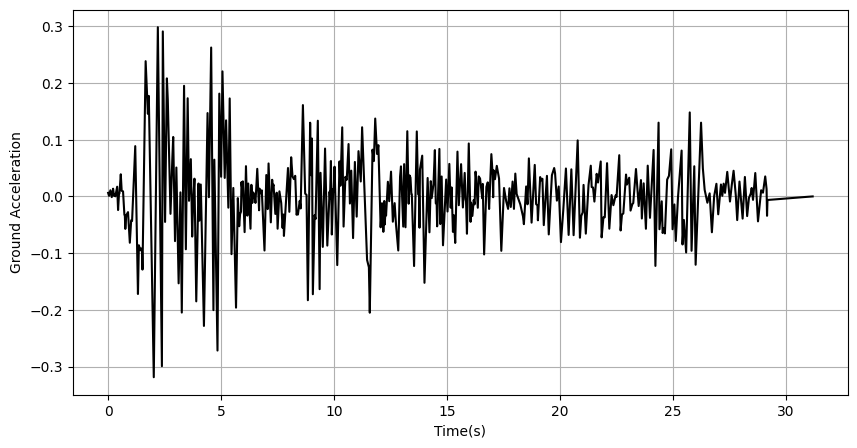} 
    \caption{Ground Acceleration time-history of Elcentro NS, 1940}
    \label{elcentro}
\end{figure}
\begin{table}[!ht]
\centering
\caption{System Parameters for Base-Excited SDoF Model}
\begin{tabular}{|c|c|c|}
\hline
\textbf{Parameter} & \textbf{Symbol} & \textbf{Value} \\ \hline
Mass & $m$ & 3.531117 kg\\ \hline
Stiffness & $k$ & 521.429791 N/m \\ \hline
Damping Ratio & $\zeta$ & 0.093387 \\ \hline
Damping Coefficient & $c = 2\zeta\sqrt{km}$ & 15.651 N.s/m\\ \hline
Time Step & $\Delta t$ & 0.02 s \\ \hline
Initial Displacement & $u_0$ & 0 \\ \hline
Initial Velocity & $v_0$ & 0 \\ \hline
Excitation Force & $F(t) = -m \ddot{u}_g(t)$ & Derived from input acceleration \\ \hline
\end{tabular}
\end{table}

\newpage
\thispagestyle{plain}
The calculated displacement, velocity and acceleration time-history is plotted for corresponding m,k,c of the system from Table 2:
\begin{figure}[!ht]
    \centering
    \includegraphics[width=0.8\textwidth]{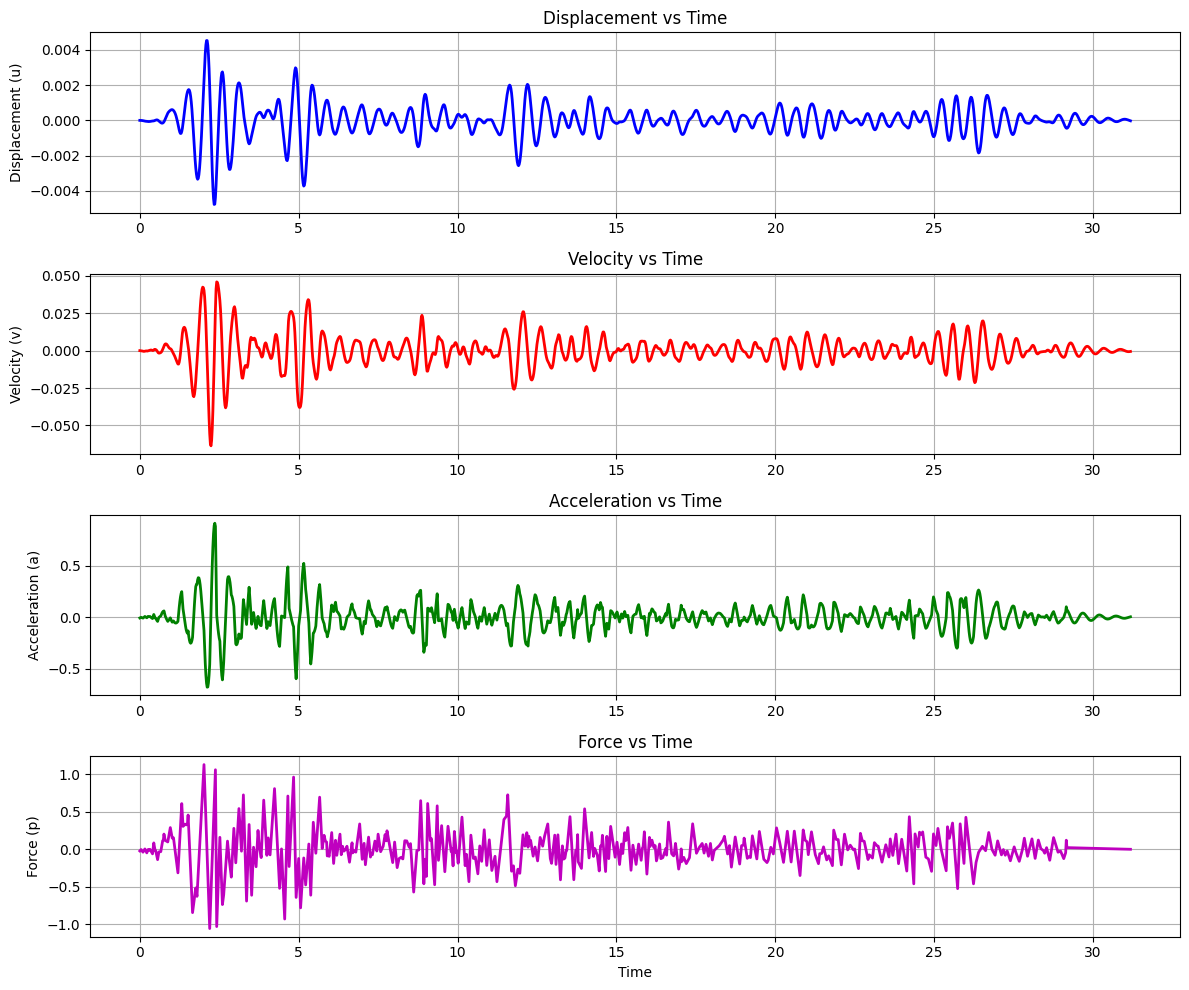} 
    \caption{Displacement, velocity, system acceleration and force time-history of a system subjected to Elcentro NS Earthquake, 1940}
    \label{response}
\end{figure}
\vspace{0.6cm}\\
\textbf{3.2 : Data Generation}\\
A numerical simulation of the SDOF system was carried out under varying mass (\(m\)), stiffness (\(k\)), and damping coefficient (\(c\)) conditions. The system response in terms of displacement time-history was obtained using the Newmark–Beta method. The maximum absolute displacement for each parameter combination was extracted to serve as the target output for the GNN model.

\vspace{0.6cm}

\textbf{3.3: Surrogate Modeling Using GNN}\\
The extracted data were used to construct a graph-based representation of the Single Degree of Freedom (SDOF) system, where nodes represented the lumped masses and edges captured the stiffness and damping relationships. The GNN was trained to learn the mapping between the structural parameters (mass 
m, damping coefficient c, and stiffness k) and the resulting maximum displacement response. The trained model was then validated against a separate set of unseen data to assess its prediction accuracy and generalization capability.

The surrogate model leverages the inherent advantages of GNNs in capturing relational dependencies through message passing between nodes and their neighbors, allowing for physically consistent modeling of the dynamic behavior of structural systems. Specifically, by encoding the mechanical system as a graph, the GNN can efficiently learn complex dependencies that are difficult to capture with traditional feedforward neural networks. Training involved minimizing a loss function measuring the deviation between predicted and simulated maximum displacements, using stochastic gradient descent and backpropagation.

\begin{figure}[!ht]
    \centering
    \includegraphics[width=1.0\textwidth]{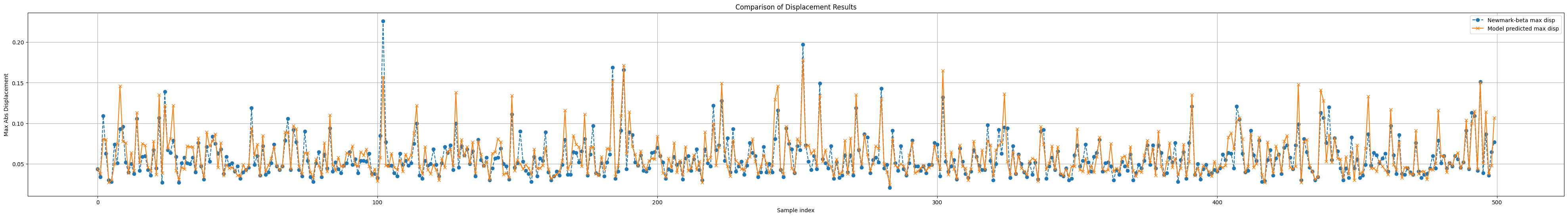} 
    \caption{GNN vs Newmark-Beta (real data) comparison plot}
    \label{nmgnn}
\end{figure}

Validation results demonstrated that the GNN surrogate achieved decently good fidelity in predicting maximum displacement across a variety of parameter combinations, significantly reducing computational cost compared to full numerical simulations. This approach enables rapid evaluations in tasks such as structural optimization, uncertainty quantification, and real-time control where repeated system analysis is required.
An outline of the current accuracy metrics of the GNN model:
\begin{table}[!ht]
\centering
\caption{Model Performance Metrics}
\begin{tabular}{|l|c|}
\hline
\textbf{Metric} & \textbf{Value} \\
\hline
$R^2$ Score & 0.7144 \\
Mean Absolute Error (MAE) & 0.009770 \\
Mean Absolute Percentage Error (MAPE) & 15.31\% \\
\hline
\end{tabular}
\end{table}\\
The model demonstrates good predictive capability with an $R^2$ score of 0.7144, indicating that approximately 71\% of the variance in maximum displacement is explained by the model. The low mean absolute error (MAE) of 0.00977 reflects high accuracy in absolute terms, while the mean absolute percentage error (MAPE) of 15.31\% indicates reasonable relative prediction error within the context of structural dynamics. Overall, the surrogate model using the graph neural network provides reliable and effective predictions of the SDOF system's maximum displacement response.
\begin{figure}[!ht]
    \centering
    \includegraphics[width=1.0\textwidth]{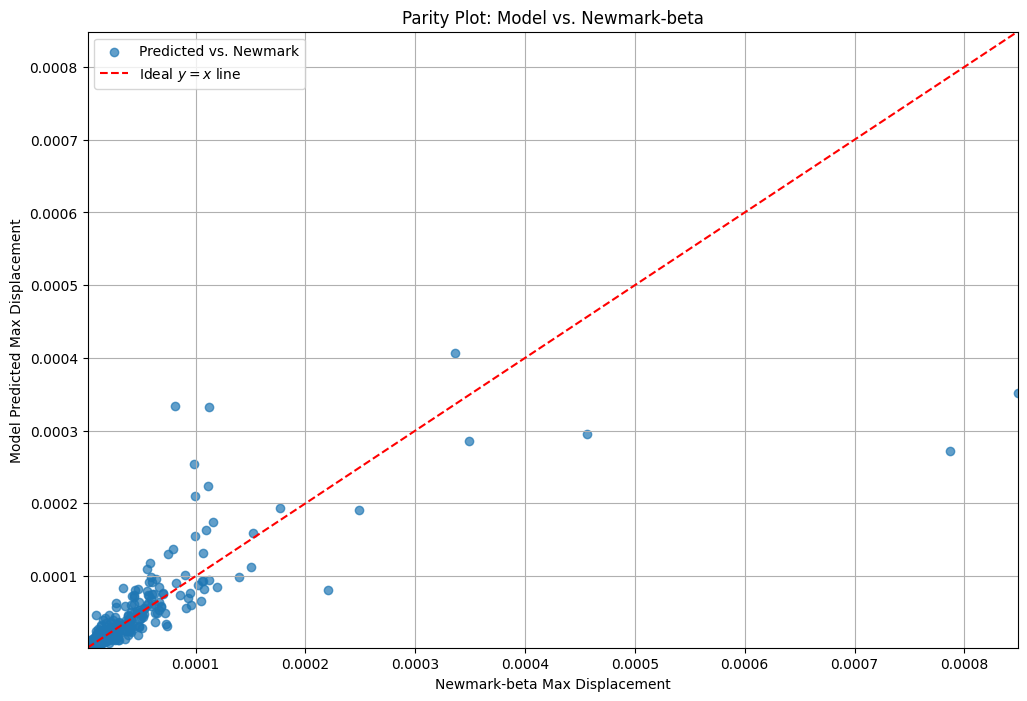} 
    \caption{Diagonal Parity Line}
    \label{dpl}
\end{figure}
\vspace{0.8cm}

\vspace{0.6cm}
\newpage
\textbf{3.4: Optimization Using Genetic Algorithm}\\
Once the surrogate model achieved satisfactory predictive accuracy, it was integrated with a Genetic Algorithm for optimization. The GA iteratively evolved candidate parameter sets through selection, crossover, and mutation operations, with the objective of minimizing the maximum displacement predicted by the GNN model.

\begin{figure}[!ht]
    \centering
    \includegraphics[width=1.0\textwidth]{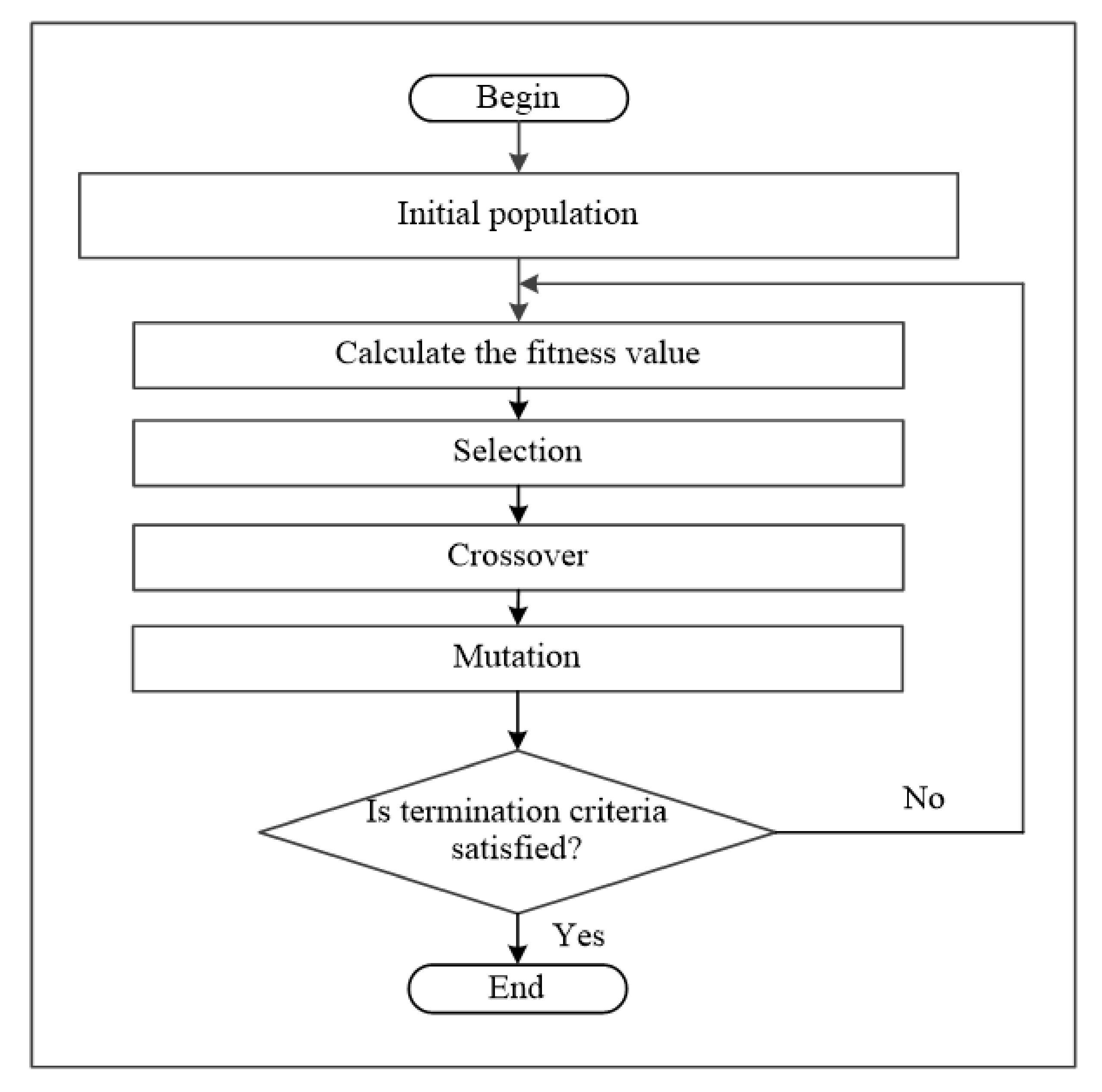} 
    \caption{Genetic Algorithm Structure}
    \label{GA_str}
\end{figure}
\begin{figure}[!ht]
    \centering
    \includegraphics[width=1.0\textwidth]{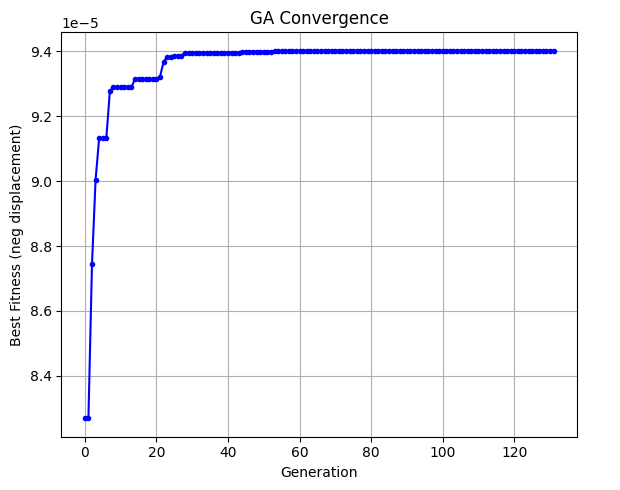} 
    \caption{Genetic Algorithm : Convergence Learning Plot}
    \label{ga_plot}
\end{figure}
\newpage
\thispagestyle{plain}
\textbf{3.5 : Validation and Comparison}\\
Finally, the optimized parameters obtained from the GA were validated through direct numerical simulation using the Newmark–Beta method. The displacement response from the simulation was compared with the GNN-predicted output to verify the effectiveness and accuracy of the proposed framework.
\begin{table}[!ht]
\centering
\caption{Comparison of Optimized Parameters and Displacement Results}
\label{tab:optimized_results}
\renewcommand{\arraystretch}{1.3} 
\begin{tabular}{|c|c|c|c|c|}
\hline
\textbf{Mass (m)} & \textbf{Stiffness (k)} & \textbf{Damping (c)} & \textbf{Displacement (GA)} & \textbf{Displacement (NM)} \\ \hline
992.2885Kg & 0.0454N/m & 0.020N.s/m &  0.000094m &  0.000097m\\ \hline

\end{tabular}
\end{table}

\section{Discussions}

The findings from the Newmark--$\beta$ solver were matched against the benchmark figures from the Average Acceleration Method to assess its precision and efficiency. Both data sets illustrate the response of a single-degree-of-freedom (SDOF) system under base excitation. The comparison involved parameters like acceleration, velocity, displacement, and external force across multiple time intervals. The strong correlation between the Newmark--$\beta$ results and the benchmark values shows that the Newmark solver effectively represents the system's dynamic behavior.

For this study, the numerical parameters selected were $\gamma = 0.5$ and $\beta = 0.25$, aligning with the average acceleration form of the Newmark method, renowned for its unconditional stability in linear systems. The current findings uphold this stability. During the time-stepping process, no numerical divergence, overshooting, or instability was noted, even with the comparatively large time step of $\Delta t = 0.02$\,s. The approach successfully preserved both numerical stability and physical consistency.

The parameters of the system analysed included $m = 3.5311~\text{kg}$, $k = 521.4298~\text{N/m}$, and $\zeta = 0.0934$. The damping coefficient was calculated to be $c = 2\zeta\sqrt{k m}$, resulting in a lightly damped system. As anticipated, the structure's vibration response demonstrated a gradual amplitude decay over time, indicative of realistic underdamped behavior. This confirms that the solver accurately models both stiffness-dominated and damping-dominated motion phases.

Slight differences between the reference and computed outcomes were noted in some displacement and velocity values. These discrepancies are primarily attributed to rounding errors, floating-point precision, and time discretization impacts in the integration process. Since the differences were negligible and did not compromise the precision or overall trend of the dynamic response, the results affirm that the developed Newmark--$\beta$ solver delivers dependable and physically coherent predictions of structural vibration.

In summary, the study proves that the implemented machine model functions accurately and remains stable under diverse excitation and parameter conditions. Its consistency with benchmark results confirms its applicability in dynamic structural analysis. Moreover, given its stability and accuracy, this model can be effectively incorporated into data-driven and optimization frameworks such as Genetic Algorithms (GA) and Graph Neural Networks (GNN) for optimizing structural parameters and predictive modeling.

\section*{References}

\begin{enumerate}
\item \label{Elcentro_NS} Chopra, A. K. \textit{Dynamics of Structures: Theory and Applications to Earthquake Engineering}, Pearson, 2020.
\item Clough, R. W., and Penzien, J. \textit{Dynamics of Structures}, Computers and Structures, 2003.
\item Nourian, N., El-Badry, M., \& Jamshidi, M. (2023). \textit{Design Optimization of Truss Structures Using a Graph Neural Network-Based Surrogate Model}. \textit{Algorithms}, 16(8), 380.
\item Chen, S. Y. and Rajan, S. D. (2000). \textit{A Robust Genetic Algorithm for Structural Optimization}. \textit{Structural Engineering \& Mechanics Journal}, Vol. 10, No. 4, pp. 313–336.
\end{enumerate}

\end{document}